\documentclass[11pt]{article}
\usepackage{bm}
\usepackage{amsmath,amsthm,verbatim,amssymb,amsfonts,amscd, graphicx,graphics,physics}
\usepackage[latin2]{inputenc}
\usepackage{t1enc}
\usepackage[mathscr]{eucal}
\usepackage{indentfirst}
\usepackage{booktabs}
\usepackage{float}
\usepackage{pict2e}
\usepackage{epic}
\numberwithin{equation}{section}
\usepackage[footnotesize,bf]{caption}
\usepackage[margin=2.9cm]{geometry}
\usepackage{epstopdf}
\usepackage{mathtools}
\usepackage[dvipsnames]{xcolor}
\usepackage{algorithmic}
\usepackage[ruled,vlined]{algorithm2e}
\usepackage{authblk}
\usepackage[titletoc]{appendix}

\usepackage{enumitem,multicol}

\usepackage{threeparttable}
\usepackage{subcaption} 
\usepackage{siunitx}
\usepackage{multirow}

\usepackage[backend=bibtex,date=year,giveninits=true,sorting=nyt,natbib=true,maxcitenames=2,maxbibnames=10,url=false,doi=true,backref=True]{biblatex}

\renewbibmacro{in:}{\ifentrytype{article}{}{\printtext{\bibstring{in}\intitlepunct}}}

\usepackage[colorlinks,linkcolor = BrickRed, citecolor=blue]{hyperref} 
\usepackage{etoolbox,cleveref}
\AtBeginEnvironment{appendices}{\crefalias{section}{appendix}}

\newlength{\leftstackrelawd}
\newlength{\leftstackrelbwd}
\def\leftstackrel#1#2{\settowidth{\leftstackrelawd}%
	{${{}^{#1}}$}\settowidth{\leftstackrelbwd}{$#2$}%
	\addtolength{\leftstackrelawd}{-\leftstackrelbwd}%
	\leavevmode\ifthenelse{\lengthtest{\leftstackrelawd>0pt}}%
	{\kern-.5\leftstackrelawd}{}\mathrel{\mathop{#2}\limits^{#1}}}

\theoremstyle{plain}
\newtheorem{theorem}{Theorem}[section]

\theoremstyle{definition}
\newtheorem{definition}[theorem]{Definition}
\newtheorem{remark}[theorem]{Remark}

\def\P{{\mathbb P}}
\def\E{{\mathbb E}}

\def\R{{\mathcal R}}
\def\X{{\mathcal X}}
\def\Y{{\mathcal Y}}
\def\H{{\mathcal H}}

\def\N{{\mathcal N}}

\def\D{{\mathcal D}}

\def\cE{{\mathcal E}}

\def\tx{{\mathbf x}}
\def\Pr{{\operatorname {Pr}}}
\def\H{{\mathcal H}}

\def\ep{{e^{\prime}}}
\def\Etr{{\cE_{\operatorname{tr}}}}
\def\Eall{{\cE_{\operatorname{all}}}}
\def\Var{{\operatorname{Var}}}

\begin{document}
	
	\title{CRIC: A robust assessment for invariant representations }
	\date{\today}
	
	
	
	\author[1]{Wenlu Tang}
	\author[2]{Zicheng Liu\footnote{Three authors contributed equally.}}
	\affil[1]{\small Department of Mathematical and Statistical Sciences, University of Alberta}
	\affil[2]{\small Department of Statistics, the Chinese University of Hong Kong, Hong Kong, China}
	\maketitle
	
	
	\vskip 0.3in
	
	
	
	
	\begin{abstract}
		The performance of machine learning models can be impacted by changes in data over time. A promising approach to address this challenge is invariant learning, with a particular focus on a method known as invariant risk minimization (IRM). This technique aims to identify a stable data representation that remains effective with out-of-distribution (OOD) data. While numerous studies have developed IRM-based methods adaptive to data augmentation scenarios, there has been limited attention on directly assessing how well these representations preserve their invariant performance under varying conditions. In our paper, we propose a novel method to evaluate invariant performance, specifically tailored for IRM-based methods. We establish a bridge between the conditional expectation of an invariant predictor across different environments through the likelihood ratio. Our proposed criterion offers a robust basis for evaluating invariant performance. We validate our approach with theoretical support and demonstrate its effectiveness through extensive numerical studies. 
		These experiments illustrate how our method can assess the invariant performance of various representation techniques.
	\end{abstract}
	
	\section{Introduction}\label{sec1}
	
	The assumption of independently and identically distributed data has been a fundamental principle in statistical machine learning.
	However, in real-world scenarios, data can originate from diverse environments, potentially violating the assumption of homogeneous distribution \cite{ahuja2021invariance, cai2023diagnosing}.
	The Empirical Risk Minimization (ERM) has been proven effective in solving this problem by considering the average loss across all training environments.
	However, if the training environments themselves exhibit shift with testing environments, the resulting model may struggle to generalize to unseen environments.
	This sensitivity to training environments introduces instability, posing challenges in applying trained models to practical scenarios.
	Consequently, it becomes crucial to explore methods that enhance generalization across potentially unseen environments, leading to the establishment of invariant learning. 
	
	The significance of invariant learning lies in its ability to enhance the adaptability and reliability of machine learning models.
	By explicitly incorporating assumptions of invariance properties \cite{peters2016causal} in training and learning invariant representations, these models can better handle scenarios where data may vary or be subject to different conditions.
	Examples of invariant learning includes {\it Domain-Adversarial Neural Network (DANN)} developed by \citet{ganin2015unsupervised} and \citet{ganin2016domain} for domain adaptation.
	\citet{li2018deep} proposed an end-to-end conditional invariant deep domain generalization approach by leveraging deep neural networks for domain-invariant representation learning.
	\citet{motiian2017unified} introduced a deep model augmenting the classification and contrastive semantic alignment loss to address the domain generalization problem.
	\citet{mitrovic2020representation} proposed {\it Representation Learning via Invariant Causal Mechanisms (RELIC)} for the classification task by enforcing the preservation of the underlying probability across different domains conditioning on an invariant representation of the covariates.
	{\it Domain-Specific Adversarial Network (DSAN)} from \citet{stojanov2021domain} offered broader applicability by assuming the invariance of the conditional distribution of the outcomes.
	\citet{arjovsky2019invariant} introduced {\it Invariant Risk Minimization (IRM)} by simply assuming the invariance of the conditional expectation of the outcome given the invariant representation.
	This approach is applicable across various machine learning tasks, encompassing classification, regression, and reinforcement learning.
	Considering the robustness of the invariance assumption and the versatility of the method, IRM emerges as a universal technique with comparatively relaxed prerequisites.
	Hence, in this paper, we will adopt IRM as the representative method for invariant learning, and subsequently present our methodology utilizing IRM as an illustrative example.
	
	IRM learns a classifier $w$ and a data representation $\Phi$ by solving the following optimization problem:
	\begin{equation}\label{sec1:IRMBasicForm}
		\min _{w, \Phi} \sum_{e\in \Etr} \mathcal{R}^e(w,\Phi)+\lambda\mathcal{J}(w,\Phi),
	\end{equation}
	where $\Etr$ is the set of all training environments.
	Here, $\mathcal{R}^e(w,\Phi)=\E_{X^e,Y^e}[l(w \circ \Phi(X^e),Y^e)]$ is the model risk under environment $e$ with $l$ being a given loss function,
	and $\mathcal{J}(w,\Phi)$ is the regularizer penalty that ensures the model to learn common features across all environments.
	Since the introduction of IRM, researchers have proposed several choices of the regularizer penalty $\mathcal{J}(w,\Phi)$.
	\citet{arjovsky2019invariant} first proposed {\it IRMv1} where $\mathcal{J}(w,\Phi)=\left\|\nabla_{w} R^e(w,\Phi)\right\|^2$.
	Under a stronger invariant assumption that $\P[Y|\Phi(X^e)]=\P[Y|\Phi(X^{\ep})]$, \citet{krueger2021out} proposed {\it REx} with $\mathcal{J}(w,\Phi)=\Var(\mathcal{R}^e(w,\Phi))$,
	while \citet{chang2020invariant} proposed {\it InvRat} with $\mathcal{J}(w,\Phi)=\lambda (R_e(w,\Phi)-R_e(w_e,\Phi))$.
	
	Many follow-up works have been proposed to improve the performance of IRM.
	\citet{rosenfeld2020risks} pointed out limitations of IRM in classification tasks.
	\citet{zhou2022sparse} added a sparsity constraint to the network and trained a neural network that is sparse to prevent overfitting.
	\citet{lin2022bayesian} updated {\it InvRat} using a Bayesian method with a posterior distribution of the classifier $w$. \citet{chang2020invariant}, \citet{koyama2020out}, \citet{ahuja2021invariance}, and \citet{li2022invariant} considered the invariant learning problem from the perspective of information theory.
	\citet{mahajan2021domain} introduced a novel regularizer to match the representation of the same object in different environments.
	\citet{creager2021environment} proposed {\it EIIL}, which attempts to automatically partition a dataset into different environments to learn environment labels that maximize the IRM's penalty.
	\citet{wang2022provable} proposed a simple post-processing method for solving the IRM problem without retraining the model.
	\citet{zhang2023federated} proposed {\it Generalization Adjustment} to address scenarios where the support of multi-domain data is not available during mini-batch training.
	
	With so many studies in the literature, however, there is currently no universally accepted criterion for evaluating the performance of an invariant representation individually.
	Traditionally, the effectiveness of invariant learning has been assessed by comparing its classification accuracy on out-of-distribution testing data to that of ERM.
	However, this comparison can be influenced by the specific data used, leading to potential biases.
	The lack of standardized measurements makes it challenging to compare the performance of different methods.
	Therefore, our study aims to address this issue by developing a robust assessment that measures the proximity of any data representation to the ideal notion of invariance, unaffected by the data normalization process.
	
	In this paper, we propose a quantity called the {\it Covariate-shift Representation Invariance Criterion (CRIC)} to serve as a robust quantifier of assessment specified above. Similar to \citet{cai2023diagnosing}, we consider to adopt likelihood ratio to quantify the shifts among various environments.
	Specifically, our approach involves utilizing the likelihood ratios of the covariates, which serve as measurements of covariate shifts, across different environments.
	We proceed by computing the variances of these ratios both after and before applying the learned data representation, and subsequently determine their quotient.
	We show that CRIC can distinguish the invariant performance of different invariant representation in our numerical studies, thus could be a good assessment for different methods that approximate an invariant representation.
	Although \citet{zhang2023nico++} introduced a metric that quantifies the presence of covariate shift and concept shift in a dataset, CRIC diverges in its primary focus, as it centers on evaluating the performance in learning an invariant representation of a method.
	
	CRIC is applicable to all methods that employ the invariance property assumption on the conditional expectation.
	To the best of our knowledge, CRIC is the first universally applicable quantity across all such methods in the existing literature.
	We highlight out methodological and theoretical contributions as follows.
	\begin{itemize}
		\item We introduce a criterion, named CRIC, to assess the invariant performance of the invariant representation obtained through invariant learning.
		This criterion is derived from the observation that the expectation of an ideal invariant predictor in one environment is equal to the expectation of the predictor weighted by a likelihood ratio in another environment.
		\item We present an empirical estimator for CRIC utilizing the available data.
		This estimator is computationally efficient and exhibits stability to linear transformations on the outcome data, making it a robust assessment method.
		\item We provide theoretical guarantees for the proposed criterion and establish the theoretical convergence of the estimator to the criterion.
		Furthermore, we conduct extensive numerical studies to provide empirical evidence supporting the theoretical findings.
		
	\end{itemize}
	The remaining paper is structured as follows.
	Section \ref{sec:bg} reviews some background knowledge about invariance property and covariate shift.
	Section \ref{sec:methodology} details the derivation of CRIC and explains how to estimate it using the available data.
	Section \ref{sec:sim} presents the experimental results that evaluate the performance of CRIC on both synthetic and real data.
	Section \ref{sec:accuracy} discusses strategies for integrating CRIC with prediction accuracy to assess the performance of invariant learning methods.
	Section \ref{sec:conclude} concludes.
	All technical proofs are deferred to the appendix.

	\section{Backgrounds}\label{sec:bg}
	
	\subsection{Invariant Risk Minimization}\label{sec:IRM}
	
	Let us consider a machine learning problem involving data that could potentially come from a collection $\Eall$ of different environments.
	Each environment is defined on the sample space $\X \times \Y$, where $\X$ and $\Y$ denote the covariate and outcome spaces, respectively.
	Denote by $X^e$ and $Y^e$ the covariate and outcome variables collected from the environment $e\in\Eall$, respectively.
	Due to sampling limitations, only data from a smaller range of environments $\Etr\subset\Eall$ can be obtained for training.
	Nevertheless, our target is to minimize the out-of-distribution risk over all possible environments, i.e.,
	\begin{equation*}
		\min_{f:\X\rightarrow\Y}R^{\operatorname{OOD}}(f)=\min_{f:\X\rightarrow\Y}\max_{e\in\Eall}R^e(f),
	\end{equation*}
	where $R^e(f)=E_{X^e,Y^e}[l(f(X^e),Y^e)]$ is the risk under environment $e$.
	The naive practice of Empirical Risk Minimization (ERM), which simply trains the model on the pooled training data, may fail to achieve this goal since the model may grasp features that only appears within $\Etr$.
	
	To tackle this issue, \citet{arjovsky2019invariant} sought to learn correlations that remain stable across different training environments, with the goal of ensuring that this stability extends to unseen environments as well.
	This transforms the main challenge into the task of finding across $\Etr$ an invariant predictor defined as follows.
	\begin{definition}\label{sec3:invarianceDefn}
		Given an embedded space $\H$, a data representation $\Phi:\X\rightarrow\H$ is said to be an {\bf invariant representation} across
		environments $\cE$ if there exists a classifier $w:\H\rightarrow\Y$ simultaneously optimal for all environments, i.e., for all $e\in\cE$,
		$$w\in\arg\max_{\bar{w}:\H\rightarrow\Y}R^e(\bar{w},\Phi).$$
		If the invariant representation $\Phi$ elicits the above classifier $w$, then $w\circ\Phi$ is called an {\bf invariant
			predictor}.
	\end{definition}
	\citet{arjovsky2019invariant} also introduced a simplified condition for the presence of an invariant predictor, specifically for common loss functions like mean squared error and cross-entropy.
	\begin{theorem}[Invariance property on expectation]\label{invRepExpVal}
		If the optimal classifier in any environment of $\cE$ can be written as a conditional expectation, then a data representation $\Phi$ is invariant if and only if, for all $e, \ep \in \cE$ and all $h$ in the intersection of the supports of $\Phi(X^e)$ and $\Phi(X^{\ep})$,
		\begin{equation}\label{invAssExp}
			\mathbb{E}[Y^e|\Phi(X^e)=h]=\mathbb{E}[Y^{\ep}|\Phi(X^{\ep})=h].
		\end{equation}
	\end{theorem}
	\begin{remark}
		Invariant learning methods universally rely on assuming certain invariance properties while conditioning on the invariant representation.
		Notably, Theorem \ref{invRepExpVal} establishes that the assumption concerning the conditional expectation \eqref{invAssExp} represents the minimal requirement.
		While some methods may adopt stronger conditions such as the invariance of the conditional distribution, the invariance on the conditional expectation remains a fundamental truth in all scenarios.
		Consequently, to ensure the broad applicability of our methodology to all invariant learning methods, we will solely assume the invariance on the conditional expectation throughout this paper.
	\end{remark}
	
	Once Definition \ref{sec3:invarianceDefn} has been established, finding the invariant predictor across $\Etr$ can be effectively reformulated as follows:
	\begin{eqnarray}
		&\min\limits_{\substack{
				\Phi:\X\rightarrow\H\\
				w:\H\rightarrow\Y
		}}&\sum_{e\in\Etr}R^e(w,\Phi),\label{sec2:IRMFormulation}\\
		&\text{subject to}& w\in\arg\min_{\bar{w}:\H\rightarrow\Y}R^e(\bar{w},\Phi),\text{ for all }e\in\Etr.\nonumber
	\end{eqnarray}
	This problem is referred to as the {\bf Invariant Risk Minimization (IRM)} problem, highlighting the significance of the invariant representation component.
	
	The IRM problem is known to be challenging to solve.
	Consequently, previous literature often approximates its solution by eliminating the constraint condition and introducing an additional term that penalizes a function measuring the degree of invariance of $\Phi$.
	This modification transforms the original formulation \eqref{sec2:IRMFormulation} into problem \eqref{sec1:IRMBasicForm}, which became the mainstream approach of IRM-based methods.
	For example, suppose that $\mathcal{D}^e=\left\{\left(\tx_i^e, \mathbf{y}_i^e\right)\right\}_{i=1}^{n_e}$ is the training dataset drawn from $e\in\Etr$ with $n_e$ being the data point size, and $\D=\{\D^e\}_{e\in\Etr}$ is the complete dataset.
	Then {\it IRMv1} \citep{arjovsky2019invariant} solves 
	\begin{equation*}
		\min_{w,\Phi} \sum_{e \in \cE_{\mathrm{tr}}} \R^e(w,\Phi)+\lambda\left\|\left.\nabla_w R^e(w,\Phi)\right|_{w=1.0}\right\|^2.
	\end{equation*}
	Section \ref{sec1} has introduced alternative choices for the penalty term $\mathcal{J}(w,\Phi)$ used in other papers.
	We encourage interested readers to refer to that section for further details.
	
	\subsection{Covariate shift and likelihood ratio}
	
	In the domains of statistical analysis and machine learning, the term {\bf covariate shift} originally refers to the situation when the distribution of training and test data are different \citep{masashi2005input}.
	In this paper, we extend the definition slightly to encompass situations where the data distributions differ across multiple environments.
	There are many covariate shift phenomena in the real-world data collected from different environments, e.g., locations, experimental conditions, times etc.
	Hence, it is not surprising that we examine the phenomenon of covariate shift within our analysis. Similar to the idea of likelihood ratio as to quantify the distribution shift in \citet{cai2023diagnosing}, we use the likelihood ratio $\rho(X^e,X^{e'})=d\P^e/d\P^{e'}$ here to reweight the expectation across different environments to achieve invariant assumption.
	
	A traditional approach to address the difference between two distributions is to reweight the distribution by the likelihood ratio to match the other. 
	For example, for two closely-related variables $X_1,X_2:\X\rightarrow\mathbb{R}$ with distributions $\P_1,\P_2$, respectively, observe that
	\begin{eqnarray*}
		\E_{X_1}(X_1) &=& \int_{\X}x\,\,d\P_1(x)=\int_{\X}x\,\,\frac{d\P_1(x)}{d\P_2(x)}\,\,d\P_2(x)\\
		&=& \E_{X_2}\left(X_2\,\,\frac{d\P_1(X_2)}{d\P_2(X_2)}\right),
	\end{eqnarray*}
	where $d\P_1(x)/d\P_2(x)$ denotes the ratio of the likelihood function, or {\bf likelihood ratio}, of $X_1$ over $X_2$.
	Thus, the likelihood ratio can be easily applied to the data that has distinct distributions across different environments.
	
	If the data contains both covariates and outcomes, previous literature may use the likelihood ratio on covariates only \citep{shimodaira2000improving, sugiyama2005input}.
	Several works used this technique of reweighting the  data samples to  adapt to target data in the covariate shift adaptation analysis \citep{sugiyama2007covariate,reddi2015doubly,chen2016robust}. 
	Existing methods usually attempt to reweight the source data samples to better represent the target domain \citep{quinonero2008dataset}.
	Previous studies have explored robust approaches for regression and conformal inference when there's a covariate shift, see \citet{chen2016robust}, \citet{tibshirani2019conformal} and \citet{candes2021conformalized}.

		\section{Methodology}\label{sec:methodology}
		In this section, we propose the robust assessment for the degree of data representation invariance, CRIC.
		We begin by deriving its analytical form based on the data distribution, and then proceed to explain how it can be estimated using the available data.
		
		\subsection{From invariance property to CRIC}
		
		Let us use the mean squared error or the cross-entropy as the loss function for a machine learning task.
		Suppose that we have a data representation $\Phi$, and we want to assess how close it is to an invariant representation on the training environment $\Etr$.
		Then by the invariance property, Theorem \ref{invRepExpVal}, we have $\mathbb{E}[Y^e|\Phi(X^e)=h]=\mathbb{E}[Y^{\ep}|\Phi(X^{\ep})=h]$ for any $e,\ep\in\Etr$ and any available $h$.
		Denoting by $\P^e$ the distribution of $X^e$, we have
		\begin{eqnarray}
			&& \E_{X^e}\left(\E(Y^e|\Phi(X^e))\right)\nonumber\\
			&=& \int_{\X}\E\left(Y^e|\Phi(X^e)=\Phi(x)\right)d\P^e(x)\nonumber\\
			&=& \int_{\X}\E\left(Y^\ep|\Phi(X^\ep)=\Phi(x)\right)d\P^e(x)\nonumber\\
			&=& \int_{\X}\E\left(Y^\ep|\Phi(X^\ep)=\Phi(x)\right)\frac{d\P^e(x)}{d\P^\ep(x)}d\P^\ep(x)\nonumber\\
			&=& \E_{X^\ep}\left(\E(Y^\ep|\Phi(X^\ep))\rho(X^e,X^\ep)(X^\ep)\right),\label{basicEqn}
		\end{eqnarray}
		where $\rho(X^e,X^\ep)=d\P^e/d\P^\ep$ is the likelihood ratio of $X^e$ to $X^\ep$.	
		
		For any $e,\ep\in\Etr$, let us denote that
		\begin{eqnarray}
			q_\Phi(e,\ep)=\E_{X^\ep}\left(\E(Y^\ep|\Phi(X^\ep))\rho(X^e,X^\ep)(X^\ep)\right).\label{qDefn}
		\end{eqnarray}
		Clearly, $q_\Phi(e,e)=\E_{X^e}\left(\E(Y^e|\Phi(X^e))\right)$. From (\ref{basicEqn}), we observe that there is only a covariate shift between $q_\Phi(e,e)$ and $q_\Phi(e,e')$.
		Thus, equation \eqref{basicEqn} suggests that any $(q_\Phi(e,\ep)-q_\Phi(e,e))^2$ should be zero when $\Phi$ is ideally invariant.
		We can then naturally argue that, for an arbitrary data representation $\Phi$, the closer $(q_\Phi(e,\ep)-q_\Phi(e,e))^2$'s are to zero, the closer $\Phi$ is to the ideal invariant representation.
		
		In order to derive a robust quantity measuring the invariance of $\Phi$ that is not affected by linear transformations of $Y$ or choices of $e,\ep$, we propose a normalized quantity,
		\begin{equation}\label{CRICDefn}
			Q_\Phi=\frac{\sum_{e,\ep\in\Etr}(q_\Phi(e,\ep)-q_\Phi(e,e))^2}{\sum_{e,\ep\in\Etr}(q_\Upsilon(e,\ep)-q_\Upsilon(e,e))^2},
		\end{equation}
		where the data representation\footnote{The predictor elicited by $\Upsilon$ is in fact the same as the predictor learned through ERM.
			Thus, the CRIC value of ERM is always $1$.} $\Upsilon$ is defined by $\Upsilon(x)=x$.
		This quantity has the following features:
		\begin{enumerate}
			\item The numerator of $Q_\Phi$ is the sum of $(q_\Phi(e,\ep)-q_\Phi(e,e))^2$ for all $e,\ep\in\Etr$.
			Therefore, it does not depend on any specific choice of environments.
			The numerator denotes the remaining variance after $\Phi$ is performed, and could reach zero if ideal.
			\item The denominator of $Q_\Phi$ is also independent of specific environments by the same reason.
			Note that it denotes the representation variance without invariant learning.
			Thus, if $Q_\Phi<1$, it indicates that $\Phi$ has acquired a certain degree of invariance across $\Etr$.
			\item If the data $Y$ is scaled by a non-zero factor $\alpha$, both the numerator and denominator of $Q_\Phi$ are scaled proportionally by $\alpha^2$.
			Therefore, the non-negative $Q_\Phi$ demonstrates robustness to linear transformations applied to the outcome variable $Y$.
		\end{enumerate}
		
		The above features have shown that $Q_\Phi$ is a good assessment quantity to the data representation invariance.
		We name this quantity the {\bf C}ovariate-shift-based {\bf R}epresentation {\bf I}nvariance {\bf C}riterion ({\bf CRIC}).
		A lower $Q_\Phi$ value naturally indicates a more invariant $\Phi$.
		For two data representations $\Phi,\Phi'$, if $Q_{\Phi}=Q_{\Phi'}$, then we consider them to have the same invariance level.
		
		\begin{remark}
			A data representation with a lower CRIC, or generally, a more invariant representation, does not necessarily guarantee a good prediction performance of the resulting elicited predictor.
			For instance, setting $\Phi(X)\equiv0$ is always an option, which yields a CRIC of $0$. However, predicting $Y$ using $\Phi(X)\equiv0$ is of limited practical utility.
			To obtain a predictor with good generalization performance across unknown environments, it is essential to strike a balance between representation invariance and training/test performance.
			While CRIC serves as a measure for the former, it is acceptable to compromise its value in order to attain better overall performance.
		\end{remark}
		
		In the remaining part of this section, we will detail the estimation procedures of CRIC.

		\subsection{Empirical estimation of CRIC}
		
		To estimate CRIC \eqref{CRICDefn} using the training dataset $\D$, we start by estimating the important intermediate term $q_\Phi(e,e')$ defined in \eqref{qDefn}.
		By Theorem \ref{invRepExpVal}, the true invariant representation $\Phi$ and the elicited classifier $w$ give the conditional expectation of $Y$.
		Thus, in theory we have
		\begin{eqnarray*}
			q_\Phi(e,\ep)=\E_{X^\ep}\left(\frac{1}{n_\ep}\sum_{i=1}^{n_\ep}w\circ\Phi(\tx_i^\ep)\rho(X^e,X^\ep)(\tx_i^\ep)\right).
		\end{eqnarray*}
		
		Suppose that we empirically learn a classifier $\hat{w}$ and data representation $\hat{\Phi}$ from $\D$.
		Then intuitively, an estimator of $q_\Phi(e,\ep)$ would be
		\begin{eqnarray}
			\hat{q}_\Phi(e,\ep)=\frac{1}{n_\ep}\sum_{i=1}^{n_\ep}\hat{w}\circ\hat{\Phi}(\tx_i^\ep)\rho(X^e,X^\ep)(\tx_i^\ep).\label{qEmperical}
		\end{eqnarray}
		
		With the general empirical estimator \eqref{qEmperical}, all other terms appearing in \eqref{CRICDefn} can be naturally estimated.
		We list them as follows.
		\begin{equation}\label{qDerivativeEmperical}
			\begin{aligned}
				\hat{q}_\Phi(e,e)&=\frac{1}{n_e}\sum_{i=1}^{n_e}\hat{w}\circ\hat{\Phi}(\tx_i^e),\\
				\hat{q}_\Upsilon(e,\ep)&=\frac{1}{n_\ep}\sum_{i=1}^{n_\ep}\tilde{w}(\tx_i^\ep)\rho(X^e,X^\ep)(\tx_i^\ep),\\
				\hat{q}_\Upsilon(e,e)&=\frac{1}{n_e}\sum_{i=1}^{n_e}\tilde{w}(\tx_i^e),
			\end{aligned}
		\end{equation}
		where $\tilde{w}$ is the naive classifier learned without employing any additional representation.
		
		Using the empirical estimations \eqref{qEmperical} and \eqref{qDerivativeEmperical}, we can estimate CRIC in the following way,
		\begin{equation}\label{CRICEmpirical}
			\hat{Q}_\Phi=\frac{\sum_{e,\ep\in\Etr}(\hat{q}_\Phi(e,\ep)-\hat{q}_\Phi(e,e))^2}{\sum_{e,\ep\in\Etr}(\hat{q}_\Upsilon(e,\ep)-\hat{q}_\Upsilon(e,e))^2}.
		\end{equation}
		
		The accuracy of CRIC estimator \eqref{CRICEmpirical} is ensured by the following theorem.
		\begin{theorem}\label{sec3:consistency}
			Suppose that, for any $e\in\Etr$, $\{\tx_i^e\}_{i=1}^{n_e}$ are independently sampled from the distribution $\P^e$, and that there exists some $a,b>0$ such that $an<n_e<bn$.
			Moreover, assume that $\Phi$ is continuous, $w$ is Lipschitz continuous and $\sum_{e,\ep\in\Etr}(q_\Upsilon(e,\ep)-q_\Upsilon(e,e))^2>0$.
			If $||\hat{\Phi}-\Phi||_\infty,||\hat{w}-w||_\infty=o(1)$ and $\tilde{w}$ is uniformly convergent as $n\rightarrow\infty$, then $|\hat{Q}_\Phi-Q_\Phi|=o_p(1)$, i.e., $|\hat{Q}_\Phi-Q_\Phi|=o(1)$ with probability $1$.
		\end{theorem}
		\begin{remark}
			The within-environment independence assumption on of $\{\tx_i^e\}_{i=1}^{n_e}$ is not a necessary condition for the convergence of CRIC's estimation.
			In fact, any assumption that enables the Law of Large Numbers to hold is acceptable, such as the common scenario of stationary time series.
			For the sake of simplicity in the proof, we assumed independence for $\{\tx_i^e\}_{i=1}^{n_e}$ in the theorem.
			
			All the other assumptions stated in Theorem \ref{sec3:consistency} are deemed reasonable.
			The assumption on the numbers of $n_e$'s guarantee that the sampling process is not imbalanced between different environments.
			For a practically applicable invariant representation, it is crucial that $\Phi$ exhibits continuity and $w$ demonstrates Lipschitz continuity in a satisfactory manner.
			The necessity of the condition $\sum_{e,\ep\in\Etr}(q_\Upsilon(e,\ep)-q_\Upsilon(e,e))^2>0$ arises from the fact that without it, there would be no need for invariant learning.
			Although the uniform convergence of $\hat{\Phi}$ and $\hat{w}$ may depend on which approximation method is used, it is satisfied for most of the existing methods in literature.
			If the impact of the outliers is negligible, $\tilde{w}$ naturally exhibits uniform convergence.
		\end{remark}
		
		\subsection{Estimation of the likelihood ratio}
		In the previous subsection, we estimated CRIC using the learned $\hat{w}$ and $\hat{\Phi}$, under the assumption of known likelihood ratio $\rho(X^e,X^\ep)$.
		Under certain assumptions regarding the distributions of $X^e$, such as the normal distribution, the likelihood functions of $X^e$ can be easily estimated by calculating the sample means and variances from $\left\{\tx_i^e\right\}_{i=1}^{n_e}$.
		However, in a more general scenario, we do not hold any prior knowledge about $\P_e$, which necessitates estimating $\rho(X^e,X^\ep)$ without assuming any specific data distributions.
		Therefore, in this subsection, we present a generic estimation strategy for the likelihood ratio, which extends the application of CRIC to a broader context.
		
		Our method mainly follows the work done by \citet{tibshirani2019conformal}.
		For two environments $e_1,e_2\in\Etr$, let $(X_{e_1,e_2},E)$ be a pair of variables that is identically distributed as $(X^e,e)|e=e_1\text{ or }e_2$.
		\citet{tibshirani2019conformal} showed that
		\begin{eqnarray*}
			\frac{\Pr(E=e_1|X_{e_1,e_2}=x)}{\Pr(E=e_2|X_{e_1,e_2}=x)}=\frac{\Pr(E=e_1)}{\Pr(E=e_2)} \frac{d\P^{e_1}(x)}{d\P^{e_2}(x)}.
		\end{eqnarray*}
		By observing that $\Pr(E=e_1|X_{e_1,e_2}=x)+\Pr(E=e_2|X_{e_1,e_2}=x)=1$, we have
		\begin{eqnarray*}
			\frac{d\P^{e_1}(x)}{d\P^{e_2}(x)}=\frac{\Pr(E=e_1|X_{e_1,e_2}=x)}{1-\Pr(E=e_1|X_{e_1,e_2}=x)}\cdot\frac{\Pr(E=e_2)}{\Pr(E=e_1)}.
		\end{eqnarray*}
		
		Apparently, $\Pr(E=e_2)/\Pr(E=e_1)$ can be estimated by $n_{e_2}/n_{e_1}$.
		On the other hand, we consider the dataset denoted by $\mathcal{C}{e_1,e_2}$, defined as follows:
		$$\mathcal{C}{e_1,e_2}={(\tx_i^e,e):e=e_1\text{ or }e_2\text{, }1\leq i\leq n_e}.$$
		We can then employ various classifiers, such as logistic regression or random forest, to estimate the conditional probability of class membership for $\mathcal{C}_{e_1,e_2}$.
		Subsequently, if $\hat{p}(x)$ represents the classifier's estimate of $\Pr(E=e_1|X_{e_1,e_2}=x)$, we can then estimate $\rho(X^{e_1},X^{e_2})(x)=d\P^{e_1}(x)/d\P^{e_2}(x)$ by
		\begin{equation}
			\hat{\rho}(X^{e_1},X^{e_2})(x)=\frac{n_{e_2}\hat{p}(x)}{n_{e_1}(1-\hat{p}(x))}.
		\end{equation}
		
		Density ratio estimation is a well-studied area, and the method described here belongs to a general class of probabilistic classification approaches.
		Two other classes of density ratio estimation methods include moment matching and the minimization of $f$-divergences, such as the Kullback-Leibler divergence.
		For a comprehensive review of these approaches and the underlying theory, we refer readers to the work of \citet{sugiyama2012destination}.
		\begin{remark}
			Similar to \citet{cai2023diagnosing}, we employs the density ratio of covariates as a means to quantify the distribution shifts encountered acrosss different environments. Our approach does not require a strict  covariate shift assumption that $P_{Y^{e'}\mid X^{e'}}=P_{Y^e\mid X^e}$. In fact, our method is flexible and can be applied to various distribution shift scenarios.
		\end{remark}

		\section{Simulation studies}\label{sec:sim}
		In this section, we apply CRIC to both synthetic and real data to assess its effectiveness.
		
		\subsection{Synthetic structural equation model data}
		
		In this subsection, we conduct experiments on a linear structural equation model (SEM) task, as previously introduced by both \citet{arjovsky2019invariant} and \citet{krueger2021out}.
		We generate the experiment dataset from $(X^e,Y^e)$ for $e\in\{0.2, 2.0, 5.0\}$.
		$X^e=(X_1^e,X_2^e)$ contains a causal effect $X_1^e$ and a non-causal effect $X_2^e$, with both $X_1$ and $X_2$ being generated as 5-dimensional vectors.
		The generation details are as follows.
		\begin{align*}
			H^e &\sim \N(0, e^2),\\
			X_1^e &\sim W_{H \rightarrow 1} H^e+\N(0, e^2),\\
			Y^e &\sim W_{1 \rightarrow Y} X_1^e+W_{H \rightarrow Y} H^e+\N(0, \sigma_y^2),\\
			X_2^e &\sim W_{Y \rightarrow 2} Y+\N(0,\sigma_2^2).
		\end{align*}
		In the context above, $W_{H \rightarrow 1},W_{1 \rightarrow Y},W_{H \rightarrow Y}$ and $W_{Y \rightarrow 2}$ are all fixed parameters.
		The value of $W_{1 \rightarrow Y}$ is consistently set to $1$ in all experiments, while the settings of the remaining parameters are varied across different experimental scenarios.
		The following illustrations depict these variations.
		\begin{enumerate}
			\item There are two settings for $W_{H \rightarrow 1},W_{H \rightarrow Y},W_{Y \rightarrow 2}$.\\
			Fully-observed (F): $W_{H \rightarrow 1},W_{H \rightarrow Y},W_{Y \rightarrow 2}=0$;\\
			Partially-observed (P): $W_{H \rightarrow 1},W_{H \rightarrow Y},W_{Y \rightarrow 2}$ are independently distributed following $\N(0,1)$.
			\item There are two settings for $\sigma_y^2,\sigma_2^2$.\\
			Homoskedastic $Y$-noise (O): $\sigma_y^2=e^2,\sigma_2^2=1$;\\
			Heteroskedastic $Y$-noise (E): $\sigma_y^2=1,\sigma_2^2=e^2$. 
		\end{enumerate}
		Thus, there are a total of four different parameter combinations: POU (hidden=1,hetero=0), PEU (hidden=1,hetero=1), FOU (hidden=0,hetero=0), and FEU (hidden=0,hetero=1), where U represents the unscrambled $(X_1,X_2)$.
		For simplicity, we will refer to them as experiments 1-4.
		
		We generate a total sample size of $n=1300$ or $1800$ independent observations from the described structural equation modeling (SEM) model.
		We allocate $800$ and $1200$ samples, respectively, from the total samples for training the IRMv1, REx-V, and ERM models, and the remaining samples are reserved for testing purposes.
		Following the training phase using various methods, we obtain corresponding estimates for the representation $\hat{\Phi}$ and the classifier $\hat{w}$ (or $\tilde{w}$ for ERM).
		These estimates are then used to calculate the CRIC value $\hat{Q}_{\Phi}$.
		The entire process is repeated 10 times for each of the settings 1-4.
		
\begin{figure}[]
			\centering 
			\includegraphics[width=0.45\textwidth, height=2.5 in]{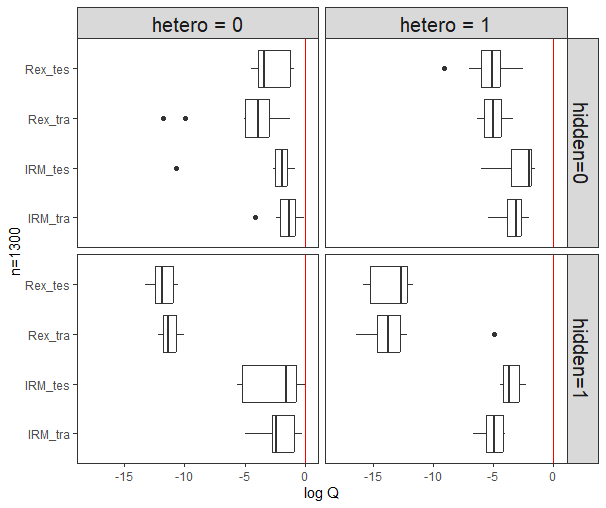}		
			\caption{
				The training and testing sample size is $800$ and $500$ respectively. The $y$-axis presents  methods IRMv1 and REx-V on testing and training data.  The red line is the baseline $\log(\hat{Q}_{\Phi})=0$ of ERM.}
			\label{Fig1} 	
		\end{figure}

\begin{figure}[]
			\center
			\includegraphics[width=0.45\textwidth, height=2.5 in]{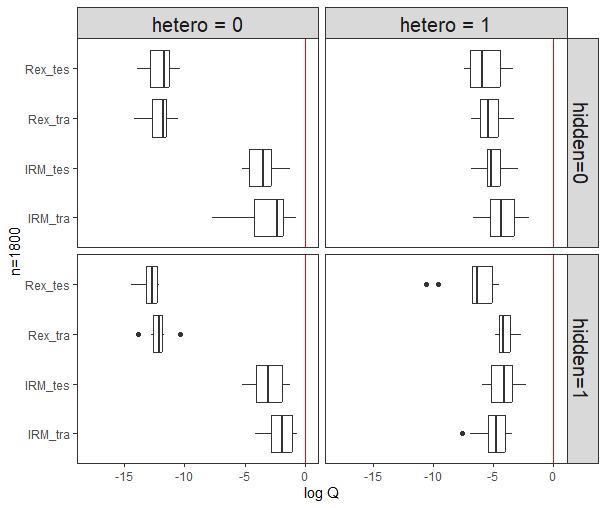}
			\caption{
				The training and testing sample size is $1200$ and $600$ respectively. The $y$-axis presents   methods IRMv1 and REx-V on testing and training data. The red line is the baseline $\log(\hat{Q}_{\Phi})=0$ of ERM.}
			\label{Fig2}
		\end{figure}

		For clarity of illustration, we present our results using $\log_{10}\hat{Q}_{\Phi}$ for both training and testing data in Figures \ref{Fig1} and \ref{Fig2}.
		The values of $\log_{10}\hat{Q}_{\Phi}$ for both IRMv1 and REx-V are consistently below $0$ in all settings, demonstrating their superiority over ERM in terms of invariant learning.
		This observation is reasonable since ERM does not capture the invariant patterns.
		In the simplest FOU and FEU settings, the performances of IRMv1 and REx-V are comparable.
		However, in the more complex setting that involves partially observed elements, REx-V exhibits better invariance performance due to its stronger penalty on the cross-environmental variance.
		
		Besides, we compare  the  estimated $\hat{q}_{\Phi}$ with true $\hat{q}_{\Phi}$ under different settings in order to show the estimation performance of each term in CRIC. In our synthetic data generation settings, the true value ${q}_\Phi(e,e^{\prime})$ are $0$ in four settings. The comparison shown in Table \ref{tab1}.
		
		
		

		
		
		\subsection{Real data}
		In this section, we utilize the IRMv1, REx-V, and ERM methods to analyze a set of financial data\footnote{https://www.kaggle.com/.},
		and then compare the CRIC values $\hat{Q}_{\Phi}$ between the IRMv1 and REx-V methods.
		We use the neural network setting in \citet{krueger2021out} and clean data consisting of factors in the U.S. stock market over five years.
		The data cleaning results in 37 features of company basic information and a target variable representing the variation of stock prices.
		The training data includes stock data from 2014 to 2016, while the testing data includes stock data from 2017 to 2018, with each year treated as an environment and varying sample sizes in each year. 
		
		Out results are summarized in Table \ref{real}.
		Again, both IRMv1 and REx-V have proven to be effective in invariant learning compared to ERM, as evidenced by all CRIC values being below $1$.
		The lower values observed for REx-V indicate that it is a more suitable method for this dataset compared to IRMv1.
		
			\begin{table}[H]
				\caption{The estimated $\hat{q}_{\Phi}$  under different settings where true  $\hat{q}_{\Phi}=0$.}
				\label{tab1}
				\centering
				\begin{tabular}{c|cc}
					\toprule
					Method   &  IRMv1 & REx-V\\
					\midrule
					POU   &$-0.079(1.122)$ & $ 0.034( 0.960)$     \\
					FOU  & $0.017 (0.576) $ & $-0.088(0.611)$  \\
					PEU &  $0.639(1.937)$ & $0.205(1.306)$     \\
					FEU  &$0.336 (1.443)$ & $-0.073(0.970)$  \\
					\bottomrule
				\end{tabular}
			\end{table}	
			
			\begin{table}[H]
				\caption{The CRIC Value of different representation methods on training ($\hat{Q}_{\Phi}$) and testing ($\hat{Q}_{\Phi}^t$) financial data.}
				\label{real}
				\centering
				\begin{tabular}{c|cc}
					\toprule
					Method     &  IRMv1 & REx-V\\
					\midrule
					$\hat{Q}_{\Phi}$ &  $0.168$ & $0.021$     \\
					$\hat{Q}_{\Phi}^t$  & $0.364 $ & $0.135$  \\
					\bottomrule
				\end{tabular}
			\end{table}
		
		\section{Integrating CRIC with Prediction Accuracy}\label{sec:accuracy}
		
		While CRIC serves as an effective measure of invariant representation invariance, it does not directly correlate with the prediction performance of the employed invariant learning method.
		CRIC is not designed to replace the prediction performance in any means, but rather as an additional criteria to emphasize the major contribution of invariance learning, i.e., increasing the robustness across environmental changes.
		Many of the recent papers are struggling in emphasizing the benefits of invariance learning methods using the prediction performance, where often the improvement is scarce.
		In this scenario, CRIC serves as an exemplary criteria evaluating the robustness across different environments, allowing researchers to shift their focus from an endless pursuit of prediction improvement and instead prioritize the assessment of overall performance and stability.
		When the prediction performance on a limited test dataset is similar among different methods, CRIC can be utilized to demonstrate the superiority of invariant learning.
		Even if there is a slight decrease in prediction performance, a substantial improvement in CRIC can still be considered a significant advancement for an algorithm.
		
		We have explored several approaches to combine CRIC with prediction performance.
		For instance, one approach involves introducing a coefficient $\theta$ to $\hat{Q}_\Phi$ and adding it to the prediction error on the test dataset, $p(X{\operatorname{tr}},Y_{\operatorname{tr}})$, resulting in an integrated criterion,
		$$p(X_,Y)+\theta \hat{Q}_\Phi.$$
		If $\theta$ is correctly chosen, a method exhibiting a smaller value of this criterion can be considered a superior choice.
		However, the value of CRIC exhibits significant variation across different types of datasets, making it impractical to establish a universal standard for selecting the coefficient.
		Ultimately, striking a balance between CRIC and prediction performance remains a matter of human preference.
		
		Given the relatively independent nature of CRIC and prediction accuracy, it is advisable to treat CRIC as a separate objective function distinct from the prediction error.
		To this extent, it is feasible to formulate the invariant learning method as a multi-objective optimization problem, where both $p(X,Y)$ and $\hat{Q}_\Phi$ are aimed to be minimized.
		As any solution located on the Pareto front can be considered acceptable to a certain extent, researchers have the flexibility to explore the most invariant representation without compromising prediction accuracy, as well as pursue the most accurate predictor without violating the invariant property.
		We leave the task of balancing between CRIC and prediction accuracy as a potential area for future investigation.
		
		\section{Concluding remarks}\label{sec:conclude}
		
		This paper introduced the Covariate-shift Representation Invariance Criterion (CRIC), which quantifies the invariance performance of data representation across potentially unseen environments in machine learning tasks, specifically for IRM-based methods.
		CRIC is based on reweighting the covariate shift between different environments using the likelihood ratio.
		It has been shown to be robust against linear transformations on the data.
		To enhance its applicability, we proposed consistent estimates for both CRIC and the likelihood ratio.
		Analyses on both synthetic and real data was conducted to demonstrate the effectiveness of CRIC, which showcased its utility in assessing data representation invariance.
		
		We acknowledge that that when selecting the optimal representation, CRIC should be considered alongside other criteria related to model accuracy.
		In future research, the emphasis will be on achieving a well-balanced combination of CRIC and performance on training/test data, enabling a comprehensive evaluation of both invariance and prediction capabilities.
		Additionally, there is potential for exploring the application of CRIC to diverse types of data, which would open up new avenues for its utilization and broaden its scope.

		
		

		\section*{Impact Statement}
		By addressing the critical challenge of evaluation of methods in domain generalization, our research not only advances the field of machine learning but also has far-reaching impacts on how to assess the predictive models developed and applied in various real-world scenarios, making systems more reliable, equitable, and adaptable to new challenges.

		\printbibliography 

@article{peters2016causal,
  title={Causal inference by using invariant prediction: identification and confidence intervals},
  author={Peters, Jonas and B{\"u}hlmann, Peter and Meinshausen, Nicolai},
  journal={Journal of the Royal Statistical Society Series B: Statistical Methodology},
  volume={78},
  number={5},
  pages={947--1012},
  year={2016},
  publisher={Oxford University Press}
}

@article{cai2023diagnosing,
  title={Diagnosing model performance under distribution shift},
  author={Cai, Tiffany Tianhui and Namkoong, Hongseok and Yadlowsky, Steve},
  journal={arXiv preprint arXiv:2303.02011},
  year={2023}
}

@article{arjovsky2019invariant,
  title={Invariant risk minimization},
  author={Arjovsky, Martin and Bottou, L{\'e}on and Gulrajani, Ishaan and Lopez-Paz, David},
  journal={arXiv preprint arXiv:1907.02893},
  year={2019}
}

@article{ahuja2021invariance,
  title={Invariance principle meets information bottleneck for out-of-distribution generalization},
  author={Ahuja, Kartik and Caballero, Ethan and Zhang, Dinghuai and Gagnon-Audet, Jean-Christophe and Bengio, Yoshua and Mitliagkas, Ioannis and Rish, Irina},
  journal={Advances in Neural Information Processing Systems},
  volume={34},
  pages={3438--3450},
  year={2021}
}

@inproceedings{chen2016robust,
  title={Robust covariate shift regression},
  author={Chen, Xiangli and Monfort, Mathew and Liu, Anqi and Ziebart, Brian D},
  booktitle={Artificial Intelligence and Statistics},
  pages={1270--1279},
  year={2016},
  organization={PMLR}
}

@inproceedings{chang2020invariant,
  title={Invariant rationalization},
  author={Chang, Shiyu and Zhang, Yang and Yu, Mo and Jaakkola, Tommi},
  booktitle={International Conference on Machine Learning},
  pages={1448--1458},
  year={2020},
  organization={PMLR}
}

@inproceedings{creager2021environment,
  title={Environment inference for invariant learning},
  author={Creager, Elliot and Jacobsen, J{\"o}rn-Henrik and Zemel, Richard},
  booktitle={International Conference on Machine Learning},
  pages={2189--2200},
  year={2021},
  organization={PMLR}
}

@inproceedings{krueger2021out,
  title={Out-of-distribution generalization via risk extrapolation (rex)},
  author={Krueger, David and Caballero, Ethan and Jacobsen, Joern-Henrik and Zhang, Amy and Binas, Jonathan and Zhang, Dinghuai and Le Priol, Remi and Courville, Aaron},
  booktitle={International Conference on Machine Learning},
  pages={5815--5826},
  year={2021},
  organization={PMLR}
}

@article{koyama2020out,
  title={Out-of-distribution generalization with maximal invariant predictor},
  author={Koyama, Masanori and Yamaguchi, Shoichiro},
  journal={CoRR},  
year={2020}
}

@inproceedings{mahajan2021domain,
  title={Domain generalization using causal matching},
  author={Mahajan, Divyat and Tople, Shruti and Sharma, Amit},
  booktitle={International conference on machine learning},
  pages={7313--7324},
  year={2021},
  organization={PMLR}
}

@inproceedings{li2022invariant,
  title={Invariant information bottleneck for domain generalization},
  author={Li, Bo and Shen, Yifei and Wang, Yezhen and Zhu, Wenzhen and Li, Dongsheng and Keutzer, Kurt and Zhao, Han},
  booktitle={Proceedings of the AAAI Conference on Artificial Intelligence},
  volume={36},
  number={7},
  pages={7399--7407},
  year={2022}
}

@inproceedings{lin2022bayesian,
  title={Bayesian invariant risk minimization},
  author={Lin, Yong and Dong, Hanze and Wang, Hao and Zhang, Tong},
  booktitle={Proceedings of the IEEE/CVF Conference on Computer Vision and Pattern Recognition},
  pages={16021--16030},
  year={2022}
}

@book{quinonero2008dataset,
  title={Dataset shift in machine learning},
  author={Quinonero-Candela, Joaquin and Sugiyama, Masashi and Schwaighofer, Anton and Lawrence, Neil D},
  year={2008},
  publisher={Mit Press}
}

@article{rosenfeld2020risks,
  title={The risks of invariant risk minimization},
  author={Rosenfeld, Elan and Ravikumar, Pradeep and Risteski, Andrej},
  journal={arXiv preprint arXiv:2010.05761},
  year={2020}
}

@article{tibshirani2019conformal,
  title={Conformal prediction under covariate shift},
  author={Tibshirani, Ryan J and Foygel Barber, Rina and Candes, Emmanuel and Ramdas, Aaditya},
  journal={Advances in neural information processing systems},
  volume={32},
  year={2019}
}

@article{sugiyama2005input,
  title={Input-dependent estimation of generalization error under covariate shift},
  author={Sugiyama, Masashi and M{\"u}ller, Klaus-Robert},
  year={2005},
  publisher={Oldenbourg Wissenschaftsverlag GmbH}
}

@article{sugiyama2007covariate,
  title={Covariate shift adaptation by importance weighted cross validation.},
  author={Sugiyama, Masashi and Krauledat, Matthias and M{\"u}ller, Klaus-Robert},
  journal={Journal of Machine Learning Research},
  volume={8},
  number={5},
  year={2007}
}

@inproceedings{wang2022provable,
  title={Provable domain generalization via invariant-feature subspace recovery},
  author={Wang, Haoxiang and Si, Haozhe and Li, Bo and Zhao, Han},
  booktitle={International Conference on Machine Learning},
  pages={23018--23033},
  year={2022},
  organization={PMLR}
}

@inproceedings{zhou2022sparse,
  title={Sparse invariant risk minimization},
  author={Zhou, Xiao and Lin, Yong and Zhang, Weizhong and Zhang, Tong},
  booktitle={International Conference on Machine Learning},
  pages={27222--27244},
  year={2022},
  organization={PMLR}
}

@article{candes2021conformalized,
  title={Conformalized Survival Analysis},
  author={Cand{\`e}s, Emmanuel J and Lei, Lihua and Ren, Zhimei},
  journal={arXiv preprint arXiv:2103.09763},
  year={2021}
}

@article{shimodaira2000improving,
	title={Improving predictive inference under covariate shift by weighting the log-likelihood function},
	author={Shimodaira, Hidetoshi},
	journal={Journal of statistical planning and inference},
	volume={90},
	number={2},
	pages={227--244},
	year={2000},
	publisher={Elsevier}
}

@article{masashi2005input,
	title={Input-dependent estimation of generalization error under covariate shift},
	author={Masashi, Sugiyama and Klaus-Robert, M{\"u}ller},
	journal={Statistics \& Risk Modeling},
	volume={23},
	number={4/2005},
	pages={249--279},
	year={2005},
	publisher={De Gruyter}
}

@inproceedings{reddi2015doubly,
	title={Doubly robust covariate shift correction},
	author={Reddi, Sashank and Poczos, Barnabas and Smola, Alex},
	booktitle={Proceedings of the AAAI Conference on Artificial Intelligence},
	volume={29},
	number={1},
	year={2015}
}

@article{sugiyama2012destination,
	title={Destination and route attributes associated with adults' walking: a review.},
	author={Sugiyama, Takemi and Neuhaus, Maike and Cole, Rachel and Giles-Corti, Billie and Owen, Neville},
	journal={Medicine and science in sports and exercise},
	volume={44},
	number={7},
	pages={1275--1286},
	year={2012}
}

@inproceedings{zhang2023federated,
  title={Federated domain generalization with generalization adjustment},
  author={Zhang, Ruipeng and Xu, Qinwei and Yao, Jiangchao and Zhang, Ya and Tian, Qi and Wang, Yanfeng},
  booktitle={Proceedings of the IEEE/CVF Conference on Computer Vision and Pattern Recognition},
  pages={3954--3963},
  year={2023}
}

@inproceedings{zhang2023nico++,
  title={Nico++: Towards better benchmarking for domain generalization},
  author={Zhang, Xingxuan and He, Yue and Xu, Renzhe and Yu, Han and Shen, Zheyan and Cui, Peng},
  booktitle={Proceedings of the IEEE/CVF Conference on Computer Vision and Pattern Recognition},
  pages={16036--16047},
  year={2023}
}

@inproceedings{mitrovic2020representation,
  title={Representation Learning via Invariant Causal Mechanisms},
  author={Mitrovic, Jovana and McWilliams, Brian and Walker, Jacob C and Buesing, Lars Holger and Blundell, Charles},
  booktitle={International Conference on Learning Representations},
  year={2020}
}

@article{stojanov2021domain,
  title={Domain adaptation with invariant representation learning: What transformations to learn?},
  author={Stojanov, Petar and Li, Zijian and Gong, Mingming and Cai, Ruichu and Carbonell, Jaime and Zhang, Kun},
  journal={Advances in Neural Information Processing Systems},
  volume={34},
  pages={24791--24803},
  year={2021}
}

@inproceedings{ganin2015unsupervised,
  title={Unsupervised domain adaptation by backpropagation},
  author={Ganin, Yaroslav and Lempitsky, Victor},
  booktitle={International conference on machine learning},
  pages={1180--1189},
  year={2015},
  organization={PMLR}
}

@article{ganin2016domain,
  title={Domain-adversarial training of neural networks},
  author={Ganin, Yaroslav and Ustinova, Evgeniya and Ajakan, Hana and Germain, Pascal and Larochelle, Hugo and Laviolette, Fran{\c{c}}ois and March, Mario and Lempitsky, Victor},
  journal={Journal of machine learning research},
  volume={17},
  number={59},
  pages={1--35},
  year={2016}
}

@inproceedings{li2018deep,
  title={Deep domain generalization via conditional invariant adversarial networks},
  author={Li, Ya and Tian, Xinmei and Gong, Mingming and Liu, Yajing and Liu, Tongliang and Zhang, Kun and Tao, Dacheng},
  booktitle={Proceedings of the European conference on computer vision (ECCV)},
  pages={624--639},
  year={2018}
}

@inproceedings{motiian2017unified,
  title={Unified deep supervised domain adaptation and generalization},
  author={Motiian, Saeid and Piccirilli, Marco and Adjeroh, Donald A and Doretto, Gianfranco},
  booktitle={Proceedings of the IEEE international conference on computer vision},
  pages={5715--5725},
  year={2017}
}
		

		\newpage
		\appendix
		\onecolumn
		\section{Proof of Theorem \ref{sec3:consistency}}
		
		\begin{proof}
			We begin by proving $|\hat{q}_\Phi(e,\ep)-q_\Phi(e,\ep)|=o_p(1)$ for any $e,\ep\in\Etr$.
			For ease of illustration, we denote that
			\begin{eqnarray*}
				\tilde{q}_\Phi(e,\ep)=\frac{1}{n_\ep}\sum_{i=1}^{n_\ep}w\circ\Phi(\tx_i^\ep)\rho(X^e,X^\ep)(\tx_i^\ep).
			\end{eqnarray*}
			Since $||\hat{\Phi}-\Phi||_\infty,||\hat{w}-w||_\infty=o(1)$, $\Phi$ is continuous and $w$ is Lipschitz continuous, we clearly have $||\hat{w}\circ\hat{\Phi}-w\circ\Phi||_\infty=o(1)$.
			Thus, we can derive that	
			\begin{eqnarray*}
				\left|\hat{q}_\Phi(e,\ep)-\tilde{q}_\Phi(e,\ep)\right|
				&=&\left|\frac{1}{n_\ep}\sum_{i=1}^{n_\ep}\hat{w}\circ\hat{\Phi}(\tx_i^\ep)\rho(X^e,X^\ep)(\tx_i^\ep)-\frac{1}{n_\ep}\sum_{i=1}^{n_\ep}w\circ\Phi(\tx_i^\ep)\rho(X^e,X^\ep)(\tx_i^\ep)\right|\\
				&=&\left|\frac{1}{n_\ep}\sum_{i=1}^{n_\ep}\left(\hat{w}\circ\hat{\Phi}(\tx_i^\ep)-w\circ\Phi(\tx_i^\ep)\right)\rho(X^e,X^\ep)(\tx_i^\ep)\right|\\
				&\leq&||\hat{w}\circ\hat{\Phi}-w\circ\Phi||_\infty\cdot\left|\frac{1}{n_\ep}\sum_{i=1}^{n_\ep}\rho(X^e,X^\ep)(\tx_i^\ep)\right|.
			\end{eqnarray*}
			Since $n_\ep>an$, by Law of Large Numbers, we have
			\begin{eqnarray*}
				\frac{1}{n_\ep}\sum_{i=1}^{n_\ep}\rho(X^e,X^\ep)(\tx_i^\ep)\stackrel{a.s.}{\rightarrow}\E_{X^\ep}\left(\rho(X^e,X^\ep)(\tx_i^\ep)\right)=\int_\X\frac{d\P^e(x)}{d\P^\ep(x)}d\P^\ep(x)=1.
			\end{eqnarray*}
			Therefore, with probability $1$ we have
			\begin{eqnarray*}
				\left|\hat{q}_\Phi(e,\ep)-\tilde{q}_\Phi(e,\ep)\right|\leq||\hat{w}\circ\hat{\Phi}-w\circ\Phi||_\infty\cdot1=o(1).
			\end{eqnarray*}
			On the other hand, again by Law of Large Numbers, we have
			\begin{eqnarray*}
				\tilde{q}_\Phi(e,\ep)
				=\frac{1}{n_\ep}\sum_{i=1}^{n_\ep}w\circ\Phi(\tx_i^\ep)\rho(X^e,X^\ep)(\tx_i^\ep)
				\stackrel{a.s.}{\rightarrow}\E_{X^\ep}\left(\E(Y^\ep|\Phi(X^\ep))\rho(X^e,X^\ep)(X^\ep)\right)
				=q_\Phi(e,\ep).
			\end{eqnarray*}
			Hence, it follows that
			\begin{eqnarray*}
				|\hat{q}_\Phi(e,\ep)-q_\Phi(e,\ep)|\leq\left|\hat{q}_\Phi(e,\ep)-\tilde{q}_\Phi(e,\ep)\right|+\left|\tilde{q}_\Phi(e,\ep)-q_\Phi(e,\ep)\right|=o_p(1).
			\end{eqnarray*}
			
			Next, we note that  $|\hat{q}_\Phi(e,e)-q_\Phi(e,e)|=o_p(1)$ is a direct conclusion of the first step.
			Thus, we clearly have
			\begin{eqnarray*}
				\left|\sum_{e,\ep\in\Etr}(\hat{q}_\Phi(e,\ep)-\hat{q}_\Phi(e,e))^2-\sum_{e,\ep\in\Etr}(q_\Phi(e,\ep)-q_\Phi(e,e))^2\right|=o_p(1).
			\end{eqnarray*}
			
			Following similar procedures, we can show that
			\begin{eqnarray*}
				\left|\sum_{e,\ep\in\Etr}(\hat{q}_\Upsilon(e,\ep)-\hat{q}_\Upsilon(e,e))^2-\sum_{e,\ep\in\Etr}(q_\Upsilon(e,\ep)-q_\Upsilon(e,e))^2\right|=o_p(1).
			\end{eqnarray*}
			
			
			Therefore, since $\sum_{e,\ep\in\Etr}(q_\Upsilon(e,\ep)-q_\Upsilon(e,e))^2>0$, it follows that $|\hat{Q}_\Phi-Q_\Phi|=o_p(1)$.
		\end{proof}

	\end{document}
